\documentclass[sigconf]{acmart}

\DeclareUnicodeCharacter{FF0C}{,}

\AtBeginDocument{%
  \providecommand\BibTeX{{%
    \normalfont B\kern-0.5em{\scshape i\kern-0.25em b}\kern-0.8em\TeX}}}

\usepackage{booktabs}
\usepackage{multirow}
\usepackage{threeparttable}
\usepackage{comment}
\usepackage{subcaption}

\usepackage{xspace}
\usepackage{amsmath}
\usepackage{amsfonts}
\usepackage{caption}

\newcommand{\ie}{i.e.\@\xspace}

\begin{document}

\title{A Unified Evaluation Framework for Multi-Annotator Tendency Learning}



\author{Liyun Zhang}
\affiliation{%
  \institution{GSFS, The University of Tokyo}
  \country{Japan}
}

\author{Fengkai Liu}
\affiliation{%
  \institution{IST, The University of Osaka}
  \country{Japan}
}

\author{Xuanmeng Sha}
\affiliation{%
  \institution{D3 Center, The University of Osaka}
  \country{Japan}
}

\author{Bowen Wang}
\affiliation{%
  \institution{SANKEN, The University of Osaka}
  \country{Japan}
}

\author{Hong Liu}
\affiliation{%
  \institution{SI, Xiamen University}
  \country{China}
}

\author{Zheng Lian}
\affiliation{%
  \institution{KAIUS, Tongji University}
  \country{China}
}


\begin{abstract}
Recent works have emerged in multi-annotator learning that shift focus from Consensus-oriented Learning (CoL), which aggregates multiple annotations into a single ground-truth prediction, to Individual Tendency Learning (ITL), which models annotator-specific labeling behavior patterns (i.e., tendency) to provide explanation analysis for understanding annotator decisions. However, no evaluation framework currently exists to assess whether ITL methods truly capture individual tendencies and provide meaningful behavioral explanations. To address this gap, we propose the first unified evaluation framework with two novel metrics: (1) Difference of Inter-annotator Consistency (DIC) quantifies how well models capture annotator tendencies by comparing predicted inter-annotator similarity structures with ground-truth; (2) Behavior Alignment Explainability (BAE) evaluates how well model explanations reflect annotator behavior and decision relevance by aligning explainability-derived with ground-truth labeling similarity structures via Multidimensional Scaling (MDS). Extensive experiments validate the effectiveness of our proposed evaluation framework.
\end{abstract}












\maketitle

\section{Introduction}
\label{sec:intro}

\begin{figure}[!t]
  \centering
  \includegraphics[width=1.0\linewidth]{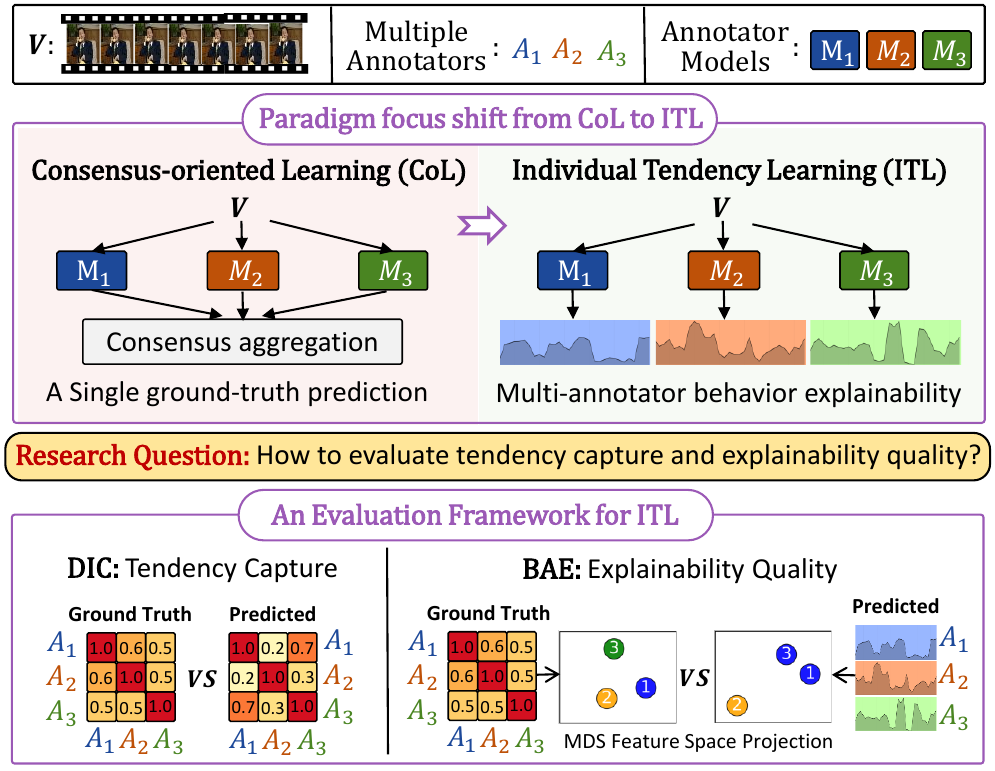}
  \caption{\textbf{Top:} A paradigm focus shift from Consensus-oriented Learning (CoL) to Individual Tendency Learning (ITL). A video sample $V$ is processed by multi-annotator models. CoL aggregates annotators' predictions into a single ground-truth prediction. While ITL models annotator-specific labeling behavior pattern (\ie, \emph{tendency}) to give different attention change explanations along video frames for understanding annotator decisions. \textbf{Bottom:} An evaluation framework for ITL: Difference of Inter-annotator Consistency (DIC) quantifies how well the model captures annotator tendencies by comparing the structure of predicted inter-annotator similarities with ground-truth; Behavior Alignment Explainability (BAE) evaluates how well model explanations reflect annotator behavior and decision relevance by aligning explainability-derived with ground-truth labeling similarity structures via Multidimensional Scaling (MDS).}
  \label{overview}
\end{figure}

In real-world multi-annotation scenarios, such as medical image analysis~\cite{PADL}, sentiment analysis~\cite{lian2023mer}, and visual perception~\cite{CNN-CM}, different annotators often provide different labels to the same sample~\cite{LFC} due to varying personal backgrounds, subjective interpretations, and expertise levels. These systematic patterns in labeling behavior—which we term \emph{tendencies}—represent valuable information about individual perspectives, cognitive biases, and domain expertise.

As illustrated in Figure~\ref{overview} (top left), the dominant paradigm in multi-annotator learning has been Consensus-oriented Learning (CoL), treating annotator disagreements as noise~\cite{noise1, Correction} to be averaged away through aggregation techniques like PADL~\cite{PADL} and MaDL~\cite{MaDL}. 
Recent works have emerged in multi-annotator learning that shift focus from CoL to Individual Tendency Learning (ITL) (Figure~\ref{overview}, top right) with sophisticated architectures like QuMAB~\cite{QuMAB} and TAX~\cite{TAX} that treat annotator diversity as valuable information to model individual annotators for capturing their unique tendencies and providing behavioral explainability.

However, despite promoting architectural innovations in ITL, there exists no principled evaluation framework to assess whether these models effectively capture annotator-specific tendencies and provide meaningful explanations to understand annotator behavior and decision relevance. This gap limits our understanding of which ITL approaches genuinely capture annotator tendencies, as opposed to those that merely optimize for consensus accuracy.

Furthermore, existing explainability assessments in multi-annotator systems rely primarily on qualitative analysis, lacking quantitative metrics to evaluate whether learned explanations accurately reflect genuine annotator behavioral relationships.
A critical question remains unanswered: How can we systematically evaluate which methods truly preserve individual annotator tendencies and provide meaningful behavioral explanations? To address these fundamental evaluation challenges, we propose a unified evaluation framework that systematically assesses ITL methods across both tendency capture and explainability quality dimensions.
Importantly, accurately reproducing individual annotations does not necessarily imply that the
relational structure within the annotator pool is preserved. Two models with similar per-annotator
accuracy may induce very different inter-annotator similarity structures, leading to divergent interpretations of group dynamics and annotator behaviors. Therefore, evaluating structural consistency among annotators provides complementary information beyond instance-level accuracy.

The \emph{Difference of Inter-annotator Consistency (DIC)} metric (Figure~\ref{overview}, bottom left) quantifies how well the model captures annotator tendencies by comparing the structure of predicted inter-annotator similarities with the ground truth. The key insight is that if an ITL method truly preserves individual tendencies, the patterns of agreement and disagreement between annotators in predictions should mirror those in actual annotations.

The \emph{Behavior Alignment Explainability (BAE)} metric (Figure~\ref{overview}, bottom right) evaluates how well model explanations reflect annotator behavior and decision relevance by aligning explainability-derived with ground-truth labeling similarity structures via Multidimensional Scaling (MDS). BAE operates through two complementary assessments: feature-level evaluation applicable to all ITL methods through learned representations, and region-level evaluation for attention-based methods through spatial/temporal attention patterns, with Figure~\ref{overview} (bottom right) demonstrating region-level evaluation. This enables systematic comparison of explainability quality across different architectural approaches.
Our work makes the following contributions:

\begin{itemize}
    \item \textbf{A unified evaluation framework for Individual Tendency Learning (ITL)}: We address the fundamental evaluation gap in ITL by introducing the first assessment framework that quantifies both tendency capture and explainability quality, enabling systematic comparison of ITL methods' effectiveness in maintaining annotator diversity while offering meaningful behavioral insights.
    
    \item \textbf{A novel metric for tendency capture evaluation}: We propose  Difference of Inter-annotator Consistency (DIC), which quantifies how well the model captures annotator tendencies by comparing the structure of predicted inter-annotator similarities with the ground truth.
    
    \item \textbf{A novel metric for explainability quality evaluation}: We propose Behavior Alignment Explainability (BAE), which evaluates how well model explanations reflect annotator behavior and decision relevance by aligning explainability-derived with ground-truth labeling similarity structures via Multidimensional Scaling (MDS).
\end{itemize}

\section{Related Work}
\label{sec:related}

\subsection{Multi-annotator Learning Paradigms}
\label{subsec:paradigms}
The dominant approach in multi-annotator learning has been consensus-oriented, aggregating diverse annotations into a single ground truth. Methods evolved from simple majority voting to sophisticated techniques using probabilistic modeling~\cite{DS_model}, EM inference~\cite{LFC, GLAD}, and deep learning~\cite{GP-MLL, Aggnet, bias_annotator}. Advanced approaches model annotator characteristics—reliability, expertise, and confusion patterns—to improve consensus quality~\cite{tanno2019learning, cao2019max, NEAL, Sampling-CM, Learn2agree}.
Recent advances explored nuanced annotator modeling: D-LEMA~\cite{D-LEMA} ensembles annotator learners, PADL~\cite{PADL} fits Gaussian distributions, and MaDL~\cite{MaDL} models confusion matrices. SimLabel~\cite{SimLabel, SimLabel1} addresses the practical challenge of missing labels in multi-annotator scenarios. Despite architectural innovations, these methods remain consensus-oriented, treating disagreements as noise to be averaged away rather than as valuable information to be modeled.
The underlying computer vision and machine learning techniques used in multi-annotator learning have broader applications across various domains~\cite{Panoptic-tcsvt, Panoptic-wacv, Panoptic1, Thermal-to-Color, PhD, 3DFacePolicy, 3DGesPolicy, uneven, Momentum, Supplementary}, though annotator disagreements in multi-annotator scenarios reflect genuine perspective differences rather than noise.
Some works recognize annotator diversity's value~\cite{CAF}, particularly in subjective domains where no absolute ground truth exists. QuMAB~\cite{QuMAB, QuMAB1} uses learnable queries to model individual patterns with interpretability. However, these efforts lack systematic evaluation frameworks to assess tendency capture and behavioral explanation quality.

\begin{figure*}[t]
    \centering
    \includegraphics[width=\textwidth]{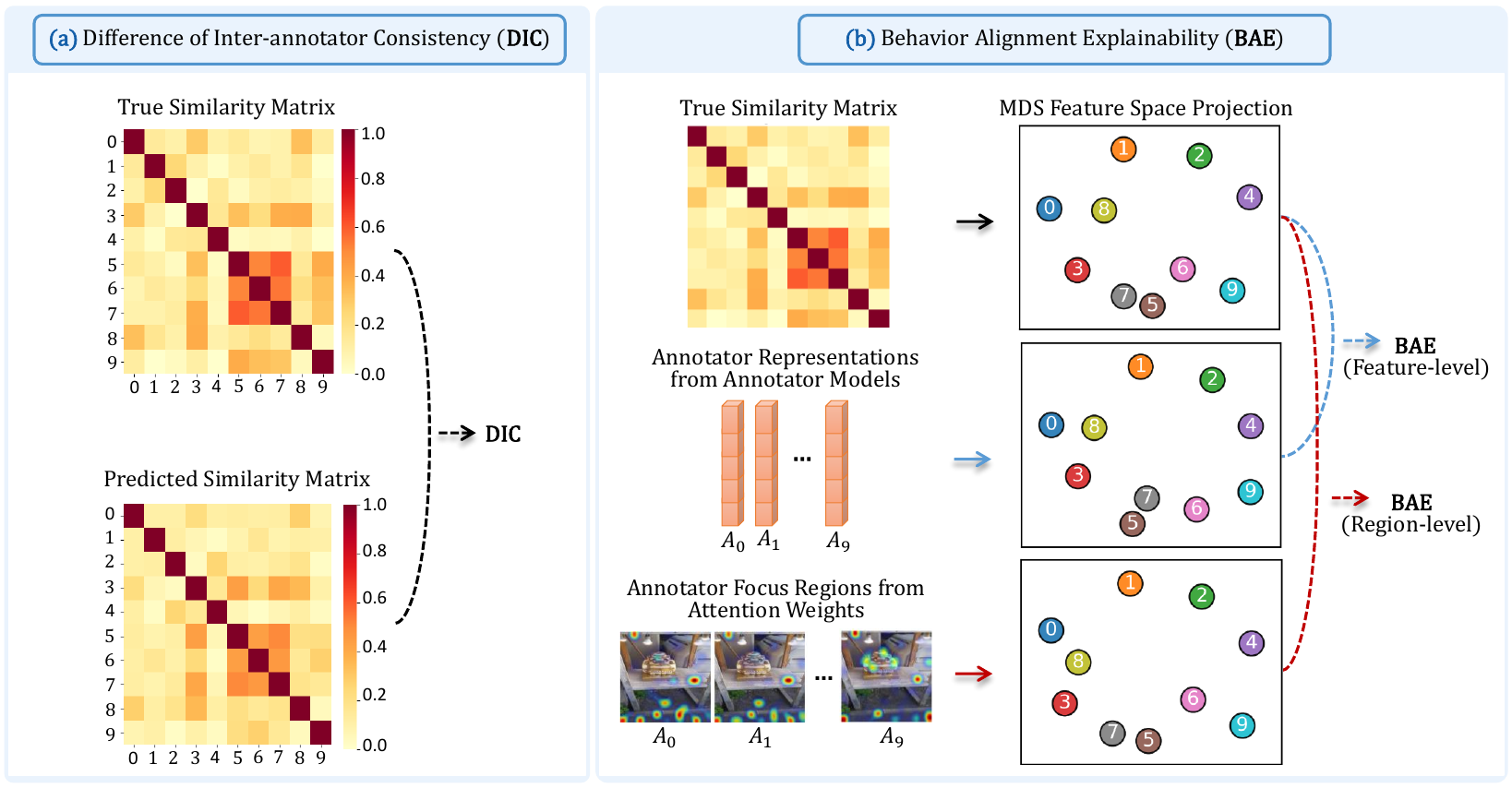}
    \caption{Proposed evaluation framework for inter-annotator behavioral analysis. 
    (a) \textbf{Difference of Inter-annotator Consistency (DIC)} quantifies how well a model preserves annotator tendency by comparing ground-truth and predicted similarity matrices using Frobenius norm. 
    (b) \textbf{Behavior Alignment Explainability (BAE)} assesses whether model explanations capture true inter-annotator behavioral structures using Multidimensional Scaling (MDS) projection. 
    BAE is computed at two complementary levels: \textit{feature-level}, based on learned annotator representations, and \textit{region-level}, based on attention-derived focus regions (for attention-based models). Both measure alignment against the ground-truth consistency matrix.}
    \label{fig:method}
\end{figure*}

\subsection{Tendency Capture Evaluation Metrics}
\label{subsec:metrics}
Evaluation in multi-annotator learning has predominantly focused on consensus accuracy, measuring how well methods predict majority votes or expert-defined ground truth~\cite{Snow2008, Carpenter2013}. Individual annotator modeling is typically assessed through prediction accuracy or log-likelihood on held-out labels~\cite{LFC, Whitehill2009}.
Traditional agreement metrics—Fleiss' kappa~\cite{Fleiss1971}, Krippendorff's alpha~\cite{KAC}, and Cohen's kappa~\cite{kappa}—measure inter-annotator agreement in raw annotations but do not assess how well computational methods preserve these patterns during learning.
Recent approaches focused on annotator modeling quality through likelihood-based metrics~\cite{Passonneau2014} or behavior correlations~\cite{Nie2020, Gordon2021}. However, these assess individual behaviors rather than inter-annotator relationship structures. Existing metrics either evaluate consensus quality or individual accuracy, but none quantify how well methods maintain inter-annotator relationships that encode perspective diversity. Our proposed Difference of Inter-annotator Consistency (DIC) metric addresses this gap by measuring tendency capture through consistency pattern matching.

\subsection{Multi-annotator Explainability Assessment}
\label{subsec:explainability}
Explainability in multi-annotator learning has received limited systematic attention, with most works providing qualitative rather than quantitative evaluation. Existing approaches include feature-based explanations using learned representations (PADL~\cite{PADL}, MaDL~\cite{MaDL}) and attention-based explanations showing spatial/temporal patterns (QuMAB~\cite{QuMAB}, TAX~\cite{TAX}), but rely on subjective interpretation without validation frameworks.
Current evaluation faces limitations: broader XAI metrics~\cite{Molnar2020, Nauta2023} target single-model scenarios and cannot assess inter-annotator relationships, while human evaluation~\cite{Doshi2017} remains expensive and difficult to standardize. Our proposed Behavior Alignment Explainability (BAE) metric addresses this gap, a quantitative assessment applicable to both explanation types. By comparing explanation-derived similarities with ground-truth behavioral patterns, BAE systematically evaluates whether explanations capture genuine annotator relationships regardless of the underlying mechanism.

\section{Methodology}
\label{sec:method}

To evaluate tendency capture and explainability, we propose two complementary metrics: Difference of Inter-annotator Consistency (DIC) and Behavior Alignment Explainability (BAE), which assess both tendency capture and explainability quality, as illustrated in Figure~\ref{fig:method}.

\subsection{Difference of Inter-annotator Consistency (DIC)}
\label{subsec:dic}

The fundamental challenge in evaluating tendency capture lies in quantifying how well a model maintains the complex web of inter-annotator relationships. We propose DIC as a principled metric that captures this preservation through consistency pattern matching.

\textbf{Core Principle.} If a model truly preserves annotator tendencies, the patterns of agreement and disagreement between annotators in predictions should mirror those in ground-truth annotations. This forms the theoretical foundation of our consistency-based evaluation approach.

\textbf{Mathematical Formulation.} Given annotations from $M$ annotators, let $Y_k = \{y_i^{(k)} : i \in \mathcal{S}_k\}$ and $Y_l = \{y_i^{(l)} : i \in \mathcal{S}_l\}$ denote the complete annotation sets for annotators $k$ and $l$, where $\mathcal{S}_k$ represents samples labeled by annotator $k$. To ensure statistical reliability, we compute consistency only on overlapping samples $\mathcal{S}_{kl} = \mathcal{S}_k \cap \mathcal{S}_l$ with $|\mathcal{S}_{kl}| \geq \tau$ (minimum threshold).

The ground-truth inter-annotator consistency matrix $\mathbf{M} \in \mathbb{R}^{M \times M}$ has elements:
\begin{align}
    m_{kl} = \kappa(Y_k|_{\mathcal{S}_{kl}}, Y_l|_{\mathcal{S}_{kl}})
\end{align}
where $\kappa(\cdot, \cdot)$ denotes Cohen's kappa coefficient. Similarly, the predicted consistency matrix $\mathbf{M}' \in \mathbb{R}^{M \times M}$ is computed as:
\begin{align}
    m'_{kl} = \kappa(\hat{Y}_k|_{\mathcal{S}_{kl}}, \hat{Y}_l|_{\mathcal{S}_{kl}})
\end{align}
where $\hat{Y}_k = \{f_{\theta}(x_i, k) : i \in \mathcal{S}_k\}$ represents predicted annotations from model $f_{\theta}$.

As illustrated in Figure~\ref{fig:method}(a), the DIC metric quantifies the preservation fidelity through direct matrix-level comparison:
\begin{align}
    \text{DIC} = \frac{\|\mathbf{M} - \mathbf{M}'\|_F}{\|\mathbf{M}\|_F}
    \label{eq:dic}
\end{align}
where normalization ensures scale-invariance across datasets. Lower DIC indicates better tendency capture, with random assignments producing high values (0.86-0.93) due to inconsistent prediction patterns, while effective methods achieve substantially lower scores by maintaining inter-annotator structural relationships. While we use Cohen’s for categorical annotations in this work, DIC only requires a well-defined inter-annotator similarity measure and can be readily extended to ordinal or continuous annotations by adopting appropriate agreement or correlation metrics.

\subsection{Behavior Alignment Explainability (BAE)}
\label{subsec:bae}

BAE assesses whether model explanations accurately capture genuine annotator behavioral relationships by evaluating the structural alignment between explanation-derived similarities and ground-truth inter-annotator consistency patterns.

\textbf{Fundamental Hypothesis.} If a model's explanations truly reflect annotator behavior patterns, the similarity relationships derived from individual annotators' representations (feature-level) or high-attention regions (region-level) on image batches or video frames should structurally align with those computed from actual annotation behaviors. We evaluate this alignment through Multidimensional Scaling (MDS) projection that enables visual comparison of similarity structures in interpretable 2D feature spaces.

\textbf{Ground-truth Behavioral Similarity.} We define the ground-truth behavioral similarity matrix $\mathbf{S}^{\text{true}} \in \mathbb{R}^{M \times M}$ where element $S^{\text{true}}_{ij}$ represents the behavioral similarity between annotators $i$ and $j$:
\begin{equation}
S^{\text{true}}_{ij} = \kappa(Y_i|_{\mathcal{S}_{ij}}, Y_j|_{\mathcal{S}_{ij}})
\end{equation}
computed over overlapping annotation sets, providing a standardized reference structure for explainability evaluation.

\textbf{Explanation-based Similarity Computation.} As shown in Figure~\ref{fig:method}(b), we compute annotator similarity matrices from model explanations at two complementary levels:

\textbf{Feature-level Assessment}: We extract annotator-specific learned representations from the penultimate layer for similarity computation:
\begin{equation}
S^{\text{feature}}_{ij} = \text{cosine}(\mathbf{F}_i^{\text{avg}}, \mathbf{F}_j^{\text{avg}})
\end{equation}
where $\mathbf{F}_i^{\text{avg}} = \frac{1}{|\mathcal{S}_i|}\sum_{x \in \mathcal{S}_i} f_{\text{feature}}(x, i)$ represents the average feature representation for annotator $i$. This assessment applies to all model architectures and evaluates how well learned representations capture behavioral distinctions.

\textbf{Region-level Assessment}: For attention-based methods, we conduct complementary region-level analysis to provide additional behavioral insights. Building upon comprehensiveness score validation~\cite{comprehensiveness}—which confirms that attention patterns focus on decision-relevant regions through performance comparison after masking high-attention versus random regions—we compute inter-annotator similarities based on attention patterns over spatial/temporal regions:
\begin{equation}
S^{\text{region}}_{ij} = \text{cosine}(\mathbf{A}_i^{\text{avg}}, \mathbf{A}_j^{\text{avg}})
\end{equation}
where $\mathbf{A}_i^{\text{avg}} = \frac{1}{|\mathcal{S}_i|}\sum_{x \in \mathcal{S}_i} \text{Attention}(x, i)$ represents the average attention pattern for annotator $i$ across input regions (image patches for STREET, video frames for AMER). This approach provides a complementary perspective by analyzing fine-grained spatial/temporal behavioral patterns, offering different analytical insights rather than consistently superior performance compared to feature-level assessment. This analysis is architecture-dependent and is not intended as a universally comparable evaluation across all ITL methods.

In practice, feature-level BAE provides a global assessment of tendency alignment and is generally
sufficient when annotator behavior differences are primarily semantic. Region-level BAE is particularly informative when annotator tendencies manifest as localized or spatially heterogeneous attention patterns, such as region-specific preferences or biases.

\textbf{Alignment Measurement.} BAE quantifies the structural alignment between explanation-derived and ground-truth behavioral similarities:
\begin{equation}
\text{BAE} = 1 - \frac{\|\mathbf{S}^{\text{model}} - \mathbf{S}^{\text{true}}\|_F}{\|\mathbf{S}^{\text{true}}\|_F}
\end{equation}
where $\mathbf{S}^{\text{model}}$ represents either feature-level or region-level similarities. Higher BAE values indicate better alignment between model explanations and true behavioral relationships. MDS projection enables interpretable visual assessment in 2D feature space, where spatial proximity reflects behavioral similarity and clustering patterns reveal whether explanations capture genuine annotator behavioral relationships.

\begin{figure*}[t]
    \centering
    \includegraphics[width=0.9\textwidth]{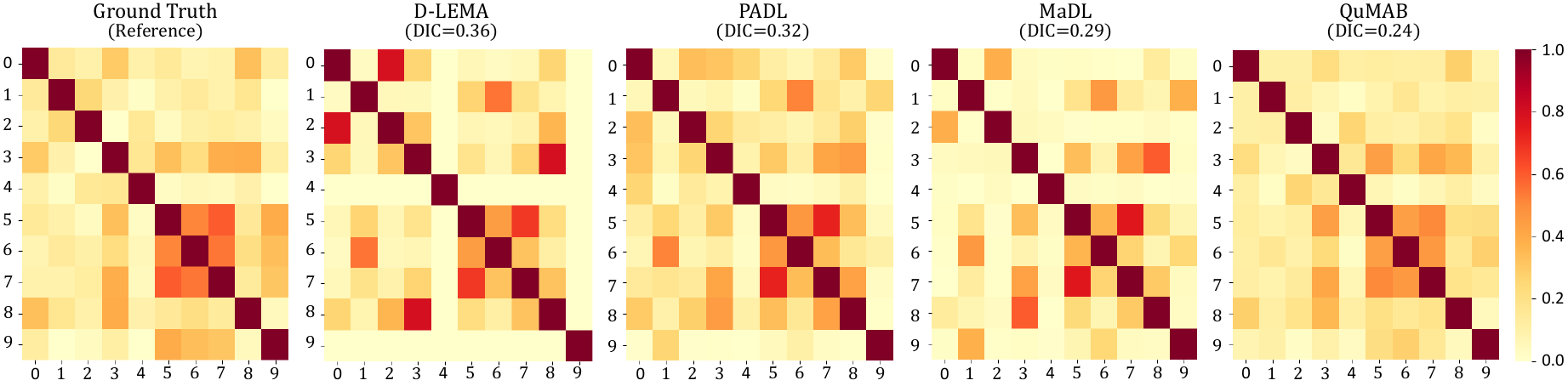}
    \caption{Visualization analysis about difference of inter-annotator consistency (DIC) via similarity matrices calculated by Cohen’s kappa coefficient on the STREET dataset (safety perspective) (10 annotators), darker colors indicate stronger agreement. Four representative models are compared with the ground truth. Lower DIC scores reflect better preservation of the underlying consistency structure. The vertical color bar denotes the similarity scale ranging from 0 (no agreement) to 1 (perfect agreement).}
    \label{fig:consistency_heatmaps}
\end{figure*}

\begin{table*}[!htbp]
\centering
\begin{tabular}{lcccccc}
\toprule
Dataset & D-LEMA & PADL & MaDL & QuMAB \\
\midrule
STREET-(Happiness) & 0.62 ± 0.03 & 0.48 ± 0.02 & 0.45 ± 0.01 & \textbf{0.43 ± 0.02} \\
STREET-(Healthiness) & 0.59 ± 0.02 & 0.52 ± 0.03 & 0.45 ± 0.03 & \textbf{0.38 ± 0.02} \\
STREET-(Safety) & 0.36 ± 0.02 & 0.32 ± 0.02 & 0.29 ± 0.01 & \textbf{0.24 ± 0.03} \\
STREET-(Liveliness) & 0.51 ± 0.01 & 0.43 ± 0.02 & 0.39 ± 0.02 & \textbf{0.27 ± 0.02} \\
STREET-(Orderliness) & 0.61 ± 0.02 & 0.57 ± 0.01 & 0.59 ± 0.02 & \textbf{0.54 ± 0.01} \\
AMER & 0.42 ± 0.03 & 0.36 ± 0.01 & 0.31 ± 0.01 & \textbf{0.23 ± 0.02} \\
\bottomrule
\end{tabular}
\caption{Difference of Inter-annotator Consistency (DIC) score comparison across different model architectures on the AMER dataset and five STREET perspectives. Lower values indicate better preservation of inter-annotator structural tendencies.}
\label{tab:consistency}
\end{table*}

\begin{table*}[!htbp]
\centering
\begin{tabular}{l@{\hspace{2.5mm}}ccccccc@{\hspace{2.5mm}}c}
\toprule
& \multicolumn{4}{c}{STREET (Safety perspective)} & \multicolumn{4}{c}{AMER} \\
\cmidrule(lr){2-5} \cmidrule(lr){6-9}
Method & ACC ↑ & FK ↑ & PCC ↑ & DIC ↓ & ACC ↑ & FK ↑ & PCC ↑ & DIC ↓ \\
\midrule
D-LEMA & 0.47 ± 0.04 & 0.51 ± 0.04 & 0.46 ± 0.05 & 0.36 ± 0.02 & 0.78 ± 0.03 & 0.54 ± 0.04 & 0.49 ± 0.05 & 0.42 ± 0.03 \\
PADL   & 0.52 ± 0.03 & 0.54 ± 0.03 & 0.49 ± 0.04 & 0.32 ± 0.02 & 0.79 ± 0.02 & 0.56 ± 0.03 & 0.52 ± 0.04 & 0.36 ± 0.01 \\
MaDL   & 0.46 ± 0.02 & 0.57 ± 0.02 & 0.52 ± 0.03 & 0.29 ± 0.01 & 0.80 ± 0.02 & 0.58 ± 0.03 & 0.55 ± 0.03 & 0.31 ± 0.01 \\
QuMAB  & \textbf{0.58 ± 0.02} & \textbf{0.61 ± 0.02} & \textbf{0.56 ± 0.02} & \textbf{0.24 ± 0.03} & \textbf{0.84 ± 0.02} & \textbf{0.63 ± 0.03} & \textbf{0.58 ± 0.03} & \textbf{0.23 ± 0.02} \\
\bottomrule
\end{tabular}
\caption{Comparison of traditional evaluation metrics and DIC on the STREET dataset (safety perspective) and the AMER dataset. While ACC, Fleiss' Kappa (FK), and Pearson Correlation Coefficient (PCC) show limited variation across methods, DIC reveals significant differences in how well models preserve annotator-specific structural tendencies.}
\label{tab:traditional_metrics}
\end{table*}

\section{Experiment}
\label{sec:experiment}
We conduct comprehensive experiments to validate our proposed evaluation framework and demonstrate its utility in assessing multi-annotator learning methods. Our evaluation focuses on two objectives: (1) validating that DIC and BAE metrics accurately reflect tendency capture and explainability quality, and (2) benchmarking representative methods to uncover insights into multi-annotator modeling capabilities.

\textbf{Implementation Details.} 
All image and video inputs are resized to 224$\times$224 and normalized before processing. Each baseline follows its original training protocol. Experiments are conducted on four NVIDIA V100 GPUs with consistent hyperparameter settings across methods.

\textbf{Datasets.} 
We evaluate on two representative multi-annotator datasets: AMER (video-based emotion recognition with 13 annotators~\cite{MicroEmo-arxiv, MicroEmo-mm}) and STREET (urban image impression assessment with 10 annotators across five dimensions: Happiness, Healthiness, Safety, Liveliness, and Orderliness), which cover complementary settings in terms of modality (video vs. image) and annotation semantics, and provide dense, diverse annotations crucial for modeling annotator-specific behavior patterns. Our focus is on evaluating Individual Tendency Learning (ITL) methods under fair and controlled conditions; thus, experiments are restricted to domains where multiple representative ITL baselines are available. Note that the proposed framework itself is modality-agnostic and can be directly extended to other domains when suitable ITL models are available.

\textbf{Baseline Methods.} 
We benchmark four representative approaches covering diverse paradigms: D-LEMA (ensemble-based aggregation), PADL (meta-learning with Gaussian fitting), MaDL (confusion-matrix-based modeling), and QuMAB (query-based annotator modeling with attention mechanisms).

\textbf{Evaluation Protocol.} Each method is evaluated using our proposed DIC and BAE metrics alongside traditional metrics to demonstrate the complementary insights provided by our framework.

\subsection{DIC Assessment}
\label{subsec:dic_evaluation}
We evaluate DIC effectiveness through visual consistency analysis, quantitative comparison across architectures, and validation against standard metrics. It is important to note that DIC evaluates the structural alignment of inter-annotator relationships, which is complementary to instance-level accuracy metrics such as ACC. A model may achieve similar DIC values while differing in ACC, reflecting different trade-offs between tendency structure modeling and pointwise prediction accuracy. Therefore, DIC is not intended to replace traditional accuracy metrics, but to be jointly considered with them for a more comprehensive evaluation.

\textbf{Consistency Pattern Analysis.} 
Figure~\ref{fig:consistency_heatmaps} presents inter-annotator consistency matrices on the STREET (safety perspective) dataset. Lower DIC scores indicate better preservation of ground-truth relationships. QuMAB achieves the most faithful reconstruction of annotator consistency patterns, while D-LEMA shows notable structural distortions. This visualization demonstrates each method's capacity to model annotator-specific relational patterns and preserve individual tendencies.

\textbf{Quantitative Results.} 
Table~\ref{tab:consistency} presents comprehensive DIC scores across four architectures on both datasets. QuMAB consistently achieves the lowest DIC scores, indicating a superior tendency to capture across visual and affective domains. Compared to D-LEMA, QuMAB improves DIC by 0.12–0.18 across perspectives, demonstrating robust capture of individualized labeling patterns under varying contexts.

\begin{figure*}[!t]
    \centering
    \includegraphics[width=\textwidth]{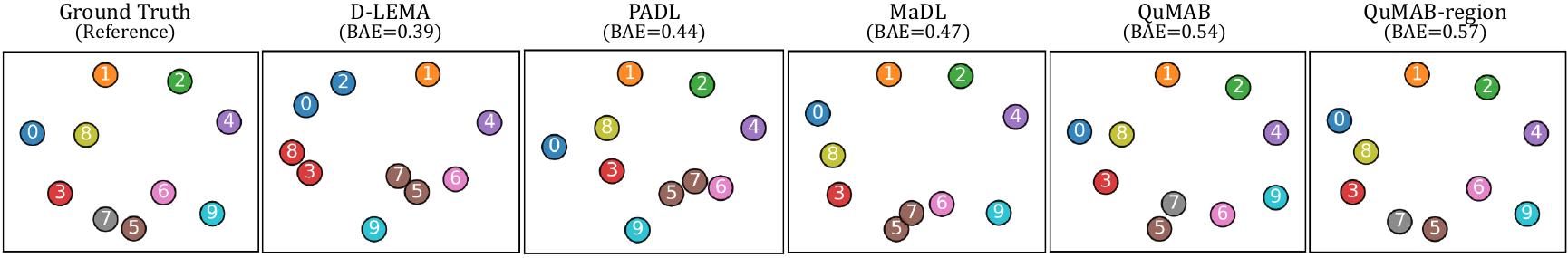}
    \caption{Visualization analysis about behavior alignment explainability (BAE) via 2D projection of annotator representations using Multidimensional Scaling (MDS) on the STREET dataset (safety perspective). Results show progressively improved alignment with higher BAE scores. QuMAB-region provides a complementary view by incorporating attention-over-region patterns. Each point denotes an annotator, and proximity indicates higher behavioral similarity. Same colors denote clusters of annotators with strong agreement ($\kappa > 0.6$).}
    \label{fig:bae_street_safety}
\end{figure*}

\begin{table*}[!t]
\centering
\begin{tabular}{lccccc}
\toprule
Dataset & D-LEMA & PADL & MaDL & QuMAB & QuMAB-region \\
\midrule
STREET-Ha & 0.28 ± 0.03 & 0.33 ± 0.04 & 0.35 ± 0.02 & 0.38 ± 0.03 & \textbf{0.41 ± 0.02} \\
STREET-He & 0.31 ± 0.04 & 0.36 ± 0.02 & 0.38 ± 0.03 & \textbf{0.42 ± 0.02} & 0.40 ± 0.03 \\
STREET-Sa & 0.39 ± 0.02 & 0.44 ± 0.03 & 0.47 ± 0.02 & 0.54 ± 0.02 & \textbf{0.57 ± 0.02} \\
STREET-Li & 0.35 ± 0.02 & 0.38 ± 0.03 & 0.41 ± 0.04 & \textbf{0.46 ± 0.02} & 0.45 ± 0.03 \\
STREET-Or & 0.24 ± 0.03 & 0.26 ± 0.02 & 0.25 ± 0.04 & 0.29 ± 0.03 & \textbf{0.31 ± 0.02} \\
AMER & 0.41 ± 0.03 & 0.45 ± 0.02 & 0.48 ± 0.01 & 0.52 ± 0.02 & \textbf{0.55 ± 0.02} \\
\bottomrule
\end{tabular}
\caption{Behavior Alignment Explainability (BAE) scores across different modeling architectures. Each value measures the alignment between learned annotator representations and ground-truth behavioral similarity. Region-level BAE (last column) evaluates inter-annotator similarity based on attention-over-region patterns mapped into feature space. Higher is better.}
\label{tab:bae_feature}
\end{table*}

\begin{table*}[!htbp]
\centering
\begin{tabular}{l@{\hspace{3.5mm}}ccc@{\hspace{3.5mm}}c@{\hspace{3.5mm}}ccc@{\hspace{3.5mm}}c}
\toprule
& \multicolumn{4}{c}{STREET (Safety perspective)} & \multicolumn{4}{c}{AMER} \\
\cmidrule(lr){2-5} \cmidrule(lr){6-9}
Method & Cos ↑ & Grad ↑ & Comp ↑ & BAE ↑ & Cos ↑ & Grad ↑ & Comp ↑ & BAE ↑ \\
\midrule
D-LEMA & 0.68 ± 0.05 & 0.54 ± 0.07 & \text{N/A} & 0.39 ± 0.02 & 0.72 ± 0.04 & 0.58 ± 0.06 & \text{N/A} & 0.41 ± 0.03 \\
PADL   & 0.71 ± 0.04 & 0.58 ± 0.06 & \text{N/A} & 0.44 ± 0.03 & 0.74 ± 0.03 & 0.61 ± 0.05 & \text{N/A} & 0.45 ± 0.02 \\
MaDL   & 0.69 ± 0.06 & 0.56 ± 0.08 & \text{N/A} & 0.47 ± 0.02 & 0.73 ± 0.05 & 0.59 ± 0.07 & \text{N/A} & 0.48 ± 0.01 \\
QuMAB  & \textbf{0.74 ± 0.03} & \textbf{0.63 ± 0.05} & 0.16 ± 0.02 & \textbf{0.54 ± 0.02} & \textbf{0.76 ± 0.02} & \textbf{0.66 ± 0.04} & 0.17 ± 0.03 & \textbf{0.52 ± 0.02} \\
\bottomrule
\end{tabular}
\caption{Comparison with alternative explainability metrics. Cos = feature cosine similarity; Grad = gradient correlation; Comp = comprehensiveness score. BAE more effectively differentiates model behavior than conventional metrics.}
\label{tab:explainability_comparison}
\end{table*}

\begin{table*}[t]
\centering
\begin{tabular}{l@{\hspace{3mm}}c@{\hspace{3mm}}c@{\hspace{3mm}}c@{\hspace{3mm}}c@{\hspace{3mm}}c@{\hspace{3mm}}c@{\hspace{3mm}}c@{\hspace{3mm}}c}
\toprule
& \multicolumn{4}{c}{DIC ↓} & \multicolumn{4}{c}{BAE ↑} \\
\cmidrule(lr){2-5} \cmidrule(lr){6-9}
Dataset & \textit{Random} & \textit{Consensus} & D-LEMA & QuMAB & \textit{Random} & \textit{Uniform} & D-LEMA & QuMAB \\
\midrule
STREET-Ha & 0.89 ± 0.03 & 0.54 ± 0.02 & 0.62 ± 0.03 & \textbf{0.43 ± 0.02} & 0.05 ± 0.04 & 0.18 ± 0.02 & 0.28 ± 0.03 & \textbf{0.38 ± 0.03} \\
STREET-He & 0.91 ± 0.04 & 0.58 ± 0.01 & 0.59 ± 0.02 & \textbf{0.38 ± 0.02} & 0.02 ± 0.03 & 0.16 ± 0.03 & 0.31 ± 0.04 & \textbf{0.42 ± 0.02} \\
STREET-Sa & 0.87 ± 0.02 & 0.51 ± 0.03 & 0.36 ± 0.02 & \textbf{0.24 ± 0.03} & 0.08 ± 0.03 & 0.23 ± 0.02 & 0.39 ± 0.02 & \textbf{0.54 ± 0.02} \\
STREET-Li & 0.90 ± 0.03 & 0.56 ± 0.01 & 0.51 ± 0.01 & \textbf{0.27 ± 0.02} & 0.04 ± 0.05 & 0.19 ± 0.01 & 0.35 ± 0.02 & \textbf{0.46 ± 0.02} \\
STREET-Or & 0.93 ± 0.05 & 0.61 ± 0.03 & 0.61 ± 0.02 & \textbf{0.54 ± 0.01} & 0.02 ± 0.06 & 0.15 ± 0.04 & 0.24 ± 0.03 & \textbf{0.29 ± 0.03} \\
AMER & 0.86 ± 0.02 & 0.53 ± 0.02 & 0.42 ± 0.03 & \textbf{0.23 ± 0.02} & 0.06 ± 0.03 & 0.21 ± 0.02 & 0.41 ± 0.03 & \textbf{0.52 ± 0.02} \\
\bottomrule
\end{tabular}
\caption{Ablation Study for DIC and BAE metrics. Left: DIC validation showing progression from Random → Consensus → D-LEMA (limit) → QuMAB (best). Right: BAE validation showing progression from Random → Uniform → D-LEMA (limit) → QuMAB (best). Both metrics demonstrate clear discriminative power across the performance spectrum.}
\label{tab:ablation_study}
\end{table*}

\textbf{Traditional Metrics Comparison.}
To further highlight the value of DIC, we compare it with three conventional metrics: (1) Accuracy (ACC) for individual annotator prediction quality; (2) Fleiss' Kappa (FK)~\cite{Fleiss} measures the overall inter-annotator agreement adjusted for chance; (3) Pearson Correlation Coefficient (PCC)~\cite{PCC} quantifies the similarity between the predicted and true annotation vectors per annotator, reflecting linear trends but not structural agreement across annotators.
Table~\ref{tab:traditional_metrics} shows results on AMER and STREET (safety perspective). While traditional metrics show limited variation (ranges within 0.06–0.09), DIC reveals more pronounced differences. D-LEMA exhibits moderate traditional scores but the highest DIC, indicating structural modeling failure. QuMAB achieves both the best traditional scores and the lowest DIC, confirming faithful modeling of individual tendencies. These findings confirm that traditional metrics may obscure structural modeling failures, whereas DIC provides clearer, complementary insights by explicitly evaluating tendency capture.

\subsection{BAE Assessment}
\label{subsec:bae_assessment}
We evaluate BAE effectiveness through visual analysis of behavioral alignment structures, quantitative comparison across feature-level with a complementary region-level perspective, and validation against alternative explainability metrics.

\textbf{Behavioral Alignment Analysis.} 
Figure~\ref{fig:bae_street_safety} shows MDS projections of learned annotator representations for STREET (safety perspective). Ground truth establishes clear behavioral clusters, with proximity indicating similarity and colors representing high-agreement groups ($\kappa > 0.6$). The progression shows: D-LEMA (0.39) with significant distortion, PADL (0.44) with moderate preservation, MaDL (0.47) with better structure, and QuMAB (0.54) with closest ground-truth alignment. QuMAB’s region-level assessment (0.57) complements feature-level analysis by projecting inter-annotator similarities—derived from high-attention input regions—into feature space, offering finer-grained behavioral insight and providing complementary fine-grained behavioral insights. These comparisons highlight each method's capacity to preserve annotator-specific behavioral relationships through learned representations.

\textbf{Quantitative Results.} 
Table~\ref{tab:bae_feature} presents comprehensive BAE scores across the four multi-annotator learning architectures for feature-level assessments and QuMAB's region-level complementary assessment. For feature-level analysis, we extract annotator-specific embeddings from the penultimate layer and compute behavioral similarities based on annotation patterns. QuMAB consistently achieves the highest feature-level BAE scores, indicating superior capture of genuine behavioral relationships. Performance improvements vary by domain: substantial gains on Safety (0.15 over D-LEMA) versus modest improvements on Orderliness (0.05), reflecting varying complexity of behavioral pattern preservation.
For QuMAB's attention-based explanations, we conduct an exploratory region-level analysis. Its scores provide complementary insights to feature-level analysis with modest and variable improvements (0.02-0.03 across most datasets). This modest enhancement does not indicate superior performance but rather reflects a different analytical perspective that captures fine-grained spatial/temporal behavioral patterns. 
The complementary nature of these two assessment levels enables more robust, comprehensive evaluation of behavioral alignment.

\textbf{Alternative Explainability Metrics Comparison.}
We also compare BAE with three existing metrics for evaluating explanation quality: (1) Feature Cosine Similarity for basic representation alignment; (2) Gradient Correlation for feature importance alignment; (3) Comprehensiveness Score validates attention faithfulness by measuring performance drops after masking high-attention regions versus random regions, applicable only to attention-based methods like QuMAB.
Table~\ref{tab:explainability_comparison} shows results on AMER and STREET (safety perspective). Additional results for the other STREET perspectives are provided in the Supplementary Material. While Feature Cosine Similarity and Gradient Correlation show consistently high scores with minimal variation (std = 0.02-0.04), BAE demonstrates substantially higher discrimination (std = 0.05-0.06), revealing clearer distinctions between methods' explainability capabilities. Traditional metrics capture surface-level similarities rather than meaningful behavioral differences.
BAE's enhanced sensitivity identifies methods with similar feature representations that fail to preserve essential behavioral relationship structures, providing complementary insights by explicitly evaluating behavioral alignment preservation across all architectural approaches.

\subsection{Ablation Study}
\label{subsec:ablation}
We conduct controlled ablation experiments using carefully designed baseline scenarios representing extreme cases of tendency capture and explainability quality to validate DIC and BAE effectiveness and sensitivity. These simulated scenarios are designed as controlled reference points to illustrate the sensitivity and discriminative behavior of the proposed metrics, rather than to serve as exhaustive
experimental benchmarks.

\textbf{DIC Ablation Analysis.} We evaluate DIC sensitivity across the performance spectrum using controlled baselines: \textit{Random} simulates completely unstructured behavior with uniformly random label assignments; \textit{Consensus} assigns majority-vote labels to all annotators, eliminating individual variations; D-LEMA provides a representative baseline with limited modeling capacity; QuMAB achieves the strongest results. As shown in the left columns of Table~\ref{tab:ablation_study} (left columns) shows DIC effectively captures the performance hierarchy: Random yields highest scores (0.86-0.93), Consensus produces moderate scores (0.51-0.61), D-LEMA shows improved performance (0.36-0.62), while QuMAB achieves lowest scores (0.23-0.54). This clear ordering validates DIC's ability to distinguish meaningful differences in tendency capture across the entire performance spectrum.

\textbf{BAE Ablation Analysis.} We validate BAE discriminative power using controlled representation scenarios: \textit{Random} generates independent random feature vectors (near-zero similarities); \textit{Uniform} assigns identical representations to all annotators (perfect similarity without behavioral differences); D-LEMA and QuMAB are selected as representative models exhibiting the lowest and highest performance, respectively. Table~\ref{tab:ablation_study} (right columns) demonstrates BAE effectiveness: Random produces the lowest scores (0.02-0.08), Uniform achieves moderate scores (0.15-0.23), D-LEMA shows improved alignment (0.24-0.41), and QuMAB reaches the highest scores (0.29-0.54). This progression confirms BAE's sensitivity to explanation quality differences and validates its utility for evaluating behavioral alignment across diverse modeling approaches.

\section{Conclusion}
\label{sec:conclusion}
We proposed the first unified evaluation framework for Individual Tendency Learning (ITL) with two novel metrics: (1) Difference of Inter-annotator Consistency (DIC) quantifies tendency capture by comparing predicted and ground-truth inter-annotator consistency structures; (2) Behavior Alignment Explainability (BAE) evaluates explainability quality through aligning explainability-derived with ground-truth labeling similarity structures via Multidimensional Scaling (MDS). Extensive experiments on different datasets across different model architectures validate the effectiveness of our proposed evaluation framework.
A potential enhancement would be incorporating human-derived behavioral signals into the evaluation framework. While eye-tracking can capture annotators’ attention patterns over image or video content, such approaches remain high-cost and difficult to scale. In future work, we aim to explore scalable alternatives for acquiring such signals, which would advance ITL and benefit the broader research community.

\bibliographystyle{ACM-Reference-Format}
\bibliography{sample-base}

\end{document}